# Enhancing Keyphrase Extraction from Microblogs using Human Reading Time

**Yingyi Zhang**

Department of Information Management, Nanjing University of Science and Technology, Nanjing, Jiangsu, China

**Chengzhi Zhang**[*]

Department of Information Management, Nanjing University of Science and Technology, Nanjing, Jiangsu, China

**Abstract**: The premise of manual keyphrase annotation is to read the corresponding content of an annotated object. Intuitively, when we read, more important words will occupy a longer reading time. Hence, by leveraging human reading time, we can find the salient words in the corresponding content. However, previous studies on keyphrase extraction ignore human reading features. In this article, we aim to leverage human reading time to extract keyphrases from microblog posts. There are two main tasks in this study. One is to determine how to measure the time spent by a human on reading a word. We use eye fixation durations extracted from an open source eye-tracking corpus (OSEC). Moreover, we propose strategies to make eye fixation duration more effective on keyphrase extraction. The other task is to determine how to integrate human reading time into keyphrase extraction models. We propose two novel neural network models. The first is a model in which the human reading time is used as the ground truth of the attention mechanism. In the second model, we use human reading time as the external feature. Quantitative and qualitative experiments show that our proposed models yield better performance than the baseline models on two microblog datasets.

**1 Introduction**

Currently, the rapid growth of user-generated content in online social media has far outpaced its capacity to be read and understood by humans. Keyphrase extraction is one of the technologies that can help organize this massive amount of content. A keyphrase consists of one or more salient words that represent the main topics of the corresponding content. It has some downstream applications in online social media, such as social influence analysis, sentiment analysis, and information retrieval. For social influence analysis, keyphrases help find topic-specific influencers on social media, which have played a crucial role in online advertising (Bi et al., 2014; Wang et al., 2014). For sentiment analysis, previous studies extracted sentiment polarity keyphrases to conduct sentiment classification (Wang et al., 2011). Moreover, keyphrases were leveraged to cluster microblog posts on the same topics, thereby assisting the retrieval task on microblogs (Choi et al., 2012).

Intuitively, when we annotate keyphrases for a document, we first need to read and understand the corresponding content. However, when we read a sentence, we do not pay the same attention to all the words. For instance, when reading the sentence '*How to integrate human attention in keyphrase extraction models*,' we may spend more time on two phrases, i.e., 'human attention' and 'keyphrase extraction models,' which are salient phrases in the above sentence. Hence, human reading time can help identify the degree of importance of a phrase. However, previous keyphrase extraction studies ignored human reading behavior and generally used the internal characteristics of the words, for example, the positions and semantic information of words ( Witten et al., 1999; Ray et al., 2019). In this study, we use human reading time in keyphrase extraction models and analyze whether it can improve their performance.

The question arises as to how to obtain the human reading time of each word. In this article, we propose using word-level eye fixation durations to determine the reading time of each word (Cop et al., 2016). The word-level eye fixation duration is the only fixations within the prescribed boundaries of each word, and it is one of the information in the eye-tracking corpus (Cop et al., 2016). The eye-tracking corpus provides human eye-movement recording while reading text, which is caught by eye trackers during human reading and watching (Reichle et al., 2003; Schiessl et al., 2003). It has been used in many fields, such as marketing research (Schiessl et al., 2003) and human-computer interaction (Strandvall, 2009). Currently, modern-day eye trackers result in rich and detailed open source eye-tracking corpora (OSECs) (Cop et al., 2016), which are easily accessible. Moreover, OSEC does not depend on target datasets and has proven efficient in various natural language processing (NLP) tasks, such as document classification (Barrett et al., 2018) and part of speech (POS) tagging (Barrett et al., 2016). Hence, we decided to obtain word-level eye fixation durations from OSECs. Generally, there are five word-level eye fixation durations measures extracted from OSECS, including first fixation duration (FFD), total reading time (TRT), single fixation duration (SFG), gaze duration (GD), and go-past time (GPT). Among them, TRT is a feature that represents the sum of all fixation durations on the current word, including regressions (Hollenstein et al., 2018). Moreover, it has been applied to various NLP tasks, such as multi-word expression prediction (Rohanian et al., 2017) and sentence classification (Barrett et al., 2018). Thus, in this study, we selected the TRT feature to represent human reading time. Table 1 presents an example in the OSECs with TRT value (Hollenstein et al., 2018).

**Table 1. An example in the OSECs with TRT value (ms)**

| Word<br>Measure | Austin | is | a | member | of | the | left-wing | Socialist | Campaign | Group. |
|---|---|---|---|---|---|---|---|---|---|---|
| **TRT** | 156 | - | - | 117 | 58 | - | 406 | 252 | 276 | 89 |

Note. "-" denotes the words were skipped during reading.

Although OSECs provide eye-tracking information, there are four major drawbacks. First, we want to know if the genre of OSEC affects the performance of keyphrase extraction on documents. To analyze this problem, we used two OSECs, the Ghent Eye-Tracking Corpus (GECO) (Cop et al., 2016) and the Zurich Cognitive Language Processing Corpus (ZUCO) (Hollenstein et al., 2018), the genres of which are 'novel' and 'Wikipedia,' respectively. Second, the eye fixation duration will be influenced by word frequency (Rayner & Duffy, 1986), which means that readers spend more time on rare words than common words. This phenomenon will have a negative impact on keyphrase extraction. For instance, in the sentence 'John Cavendish frowned and changed the subject,' theoretically the verbs 'frowned' and 'changed' are equally important. However, the eye fixation duration on 'frowned' is 324 ms, which is much longer than it is on 'changed' (lower than 100 ms). The reason is that 'frowned' is rarer than 'changed.' In the British National Corpus[1] (BNC), the word frequencies of the two words are 1312 and 11486, respectively. To mitigate the impact of word frequency on eye fixation duration, we propose a word frequency regularization method that will be introduced in Section 5.1.2. Third, previous studies showed clearly that as word length increased, the probability of fixating the word increased (Rayner & McConkie, 1976; Reichle et al., 2003). For instance, as shown in Table 1, longer words tend to have longer eye fixation duration. In this case, simply using the original eye fixation duration will lead to a misunderstanding that longer words are more important in the context. To alleviate the impact of word length on eye fixation duration, we propose a word length regularization method that will be introduced in Section 5.1.3. Fourth, the word quantity in OSEC is limited, which causes a problem in that not all words in the target dataset have corresponding eye-tracking information. As shown in Table 2, the word coverage ratio of OSEC within the two microblog datasets is low. For instance, in the training dataset of Election-Trec, the word coverage ratio of the ZUCO is 2.3%. To mitigate the limited-quantity problem, the previous study used the average of eye fixation durations to represent the human reading time of words that cannot be found in OSEC (Zhang & Zhang, 2019). In this way, different words may have the same eye fixation durations, which is an external noise to the keyphrase extraction models. To remedy this defect, we propose a similarity-based OSEC expanding strategy, which will be introduced in Section 5.1.4.

---

[1] The British National Corpus is a 100-million-word collection of samples of written and spoken language from a wide range of sources.

Table 2. Word coverage ratio (%) of OSECs within two microblog datasets

| OSEC / Microblog dataset | | GECO | | | ZUCO | | |
|---|---|---|---|---|---|---|---|
| | | Train | Vali | Test | Train | Vali | Test |
| Election-Trec | Uni | 7.8 | 18.0 | 18.1 | 2.3 | 6.0 | 5.9 |
| | All | 53.5 | 53.4 | 53.5 | 34.9 | 35.0 | 34.9 |
| General-Twitter | Uni | 4.2 | - | 6.2 | 1.3 | - | 1.9 |
| | All | 41.8 | - | 41.9 | 27.8 | - | 27.8 |

Note: **Train** denotes the dataset for model training. **Vali** denotes the validation dataset. **Test** denotes the dataset for model testing. *Uni* denotes the deduplicated dataset, which means in the dataset, the same words in different contexts are considered as one word. *All* denotes the original datasets in which the words are not duplicated.

This article aims to use human reading time in keyphrase extraction models. Hence, there is another question as to how to integrate keyphrase extraction models with human reading time. In this article, we proposed two types of neural keyphrase extraction models. In the first model, we apply an attention mechanism, in which the human reading time is regarded as the ground truth of the output of the attention mechanism. In the second model, we use human reading time as an external feature. The above-mentioned models will be described in detail in Section 3. The reason we chose neural network models is that they can automatically learn features from training samples and have proven efficient and effective on many tasks of natural language processing, for example, name entity recognition (Bahdanau et al., 2014) and event extraction (Yubo et al., 2015).

The experiments on two microblog datasets show that keyphrase extraction models with human reading time can outperform strong baselines. The analysis experiments indicate that OSECs in genres that are close to common human expression are more efficient. Moreover, the word frequency regularization method, word length regularization method, and similarity-based OSEC expanding strategy have a positive effect on the performance of keyphrase extraction.

The contribution of this article is fourfold. First, we propose two novel neural models to integrate human reading time in the keyphrase extraction task. Second, we use two OSECs to evaluate whether the genre of OSEC influences the performance of keyphrase extraction. Third, we employ a word frequency and word length regularization method to alleviate the negative impact of word frequency and word length in OSEC, respectively. Fourth, we propose a similarity-based OSEC expanding strategy to mitigate the limited-quantity problem in OSEC.

## 2 Related Work

This article aims to use human reading times collected from OSECs to facilitate keyphrase extraction on microblog posts. Hence, in this section, the first part is about previous studies on the microblog keyphrase extraction task. In the second part, besides concluding the existing OSEC and its applications in natural language processing, we introduce two OSECs that will be used in this article and the reasons for choosing them.

### 2.1 Method of Keyphrase Extraction from Microblogs

Recently, keyphrase extraction technologies have been extended to microblogs (Bellaachia & Al-Dhelaan, 2012; Zhao et al., 2011), for example, Twitter[2] and Sina Weibo[3]. Previous studies used both unsupervised models and supervised models to extract keyphrases from microblogs.

For unsupervised models, graph-based ranking models, including TextRank (Mihalcea, & Tarau,

---
[2] https://twitter.com/
[3] https://www.weibo.com/us

2004) and NE-Rank (Bellaachia & Al-Dhelaan, 2012), are frequently used. In these models, the words in the microblog are regarded as vertices. If two words have relations, there is an edge between these two words. For supervised models, recent studies have focused on selecting efficient features. For instance, Marujo et al. (2015) expanded the MAUI model with Brown clustering and word embedding features. Ray Chowdhury et al (2019) incorporate word embeddings, POS-tags, phonetics, and phonological features with models proposed by Q. Zhang et al. (2016). Y. Zhang et al. (2018) and Y. Zhang et al. (2019) encoded the conversation context consisting of replies to Tweets in neural models, and their proposed models yielded better performance than Q. Zhang et al. (2016), which proves the effectiveness of external knowledge. Hence, this article is in the line of integrating external knowledge into neural network models, and we explore the idea of using human reading times estimated from OSECs to help in keyphrase extraction.

Recently, a few studies have explored the idea of using human attention in keyphrase extraction tasks. Zhang and Zhang (2019) first proposed a human-reading-time-based keyphrase extraction model. Although this model yielded better performance than models without human reading time, there are still some drawbacks. First, they collected human reading times from one OSEC, which ignores the impact of genre on keyphrase extraction performance. Second, there are many words in target posts that cannot be found in the OSECs, which means these words do not have corresponding human reading times. Third, they did not control the word length factor in the experiment, which is also an impact factor of eye fixation duration in the OSECs. In this article, to solve the first problem, we collected the eye fixation duration from two OSECs and analyzed whether the genre of the OSEC influences performance. To solve the second problem, we propose an OSEC expanding strategy to expand the word quantity of the OSEC. To solve the third problem, .we employ a word length regulation strategy to control the impact of word length on the eye fixation duration. Moreover, we propose two types of neural models to integrate human reading time into the keyphrase extraction task.

**2.2 Open Source Eye-Tracking Datasets and Their Application in Natural Language Processing**

The open-source eye-tracking corpora include GECO (Cop et al., 2016) and ZUCO (Hollenstein et al., 2018). There are two categories of eye-tracking corpora: natural reading corpora and task-specific corpora. For task-specific corpora, participants need to answer a question after reading.

They were told the questions before reading. For natural reading corpora, to control reading quality more effectively, questions are given randomly after reading, and participants do not know these questions in advance. GECO is a natural reading corpus in which participants are required to read a novel. ZUCO has three different subcorpora. The second one is a natural reading corpus in which participants are required to read sentences from Wikipedia.

Eye-tracking information has been used in a range of natural language processing tasks, such as POS tagging and sentence classification. For POS tagging, Barrett et al. (2016) extracted human reading features from an eye-tracking corpus and combined these features with a hidden Markov model. Sentence classification tasks rely on the assumption that humans will filter less important information while reading, and eye fixation duration can indicate which portions of the text carry higher weights than others (Barrett et al., 2018).

This article tries to use GECO and the second sub-ZUCO corpus. The first reason is that they are natural reading corpora. In task-specific corpora, participants may pay more attention to words related to questions. For instance, participants will spend more time looking for sentiment polarity words in sentiment analysis tasks. Because the keyphrase extraction task is different from tasks in eye-tracking experiments, we used natural reading corpora to control the noise caused by specific tasks. What's more, the genre of GECO and second sub-ZUCO are novel and Wikipedia, respectively. With these two corpora in different genres, we can effectively analyze the impact of genre on the performance of the keyphrase extraction task.

## 3. Keyphrase Extraction Model with Human Reading Time

In this section, we propose two novel keyphrase extraction models with human reading time. The first is a neural network model with an attention mechanism, and we use the human reading time as the ground truth of the outputs of the attention mechanism. The second is a model in which the human reading time is regarded as an external feature. Sections 3.1 and 3.2 give a brief outline of these two models, respectively.

### 3.1 Attention-Mechanism-based Keyphrase Extraction Model

The attention-mechanism-based keyphrase extraction model consists of two modules: the keyphrase extraction module and the attention mechanism module. The keyphrase extraction module is used for extracting keyphrases from target microblogs. The attention mechanism module measures the degree of importance of each word in the target microblog by simulating human reading behavior (Bahdanau et al., 2014). Besides the traditional attention mechanism, this module uses human reading time as the ground truth of the output of the attention mechanism, by which the output of the attention mechanism can be revised by the human reading time.

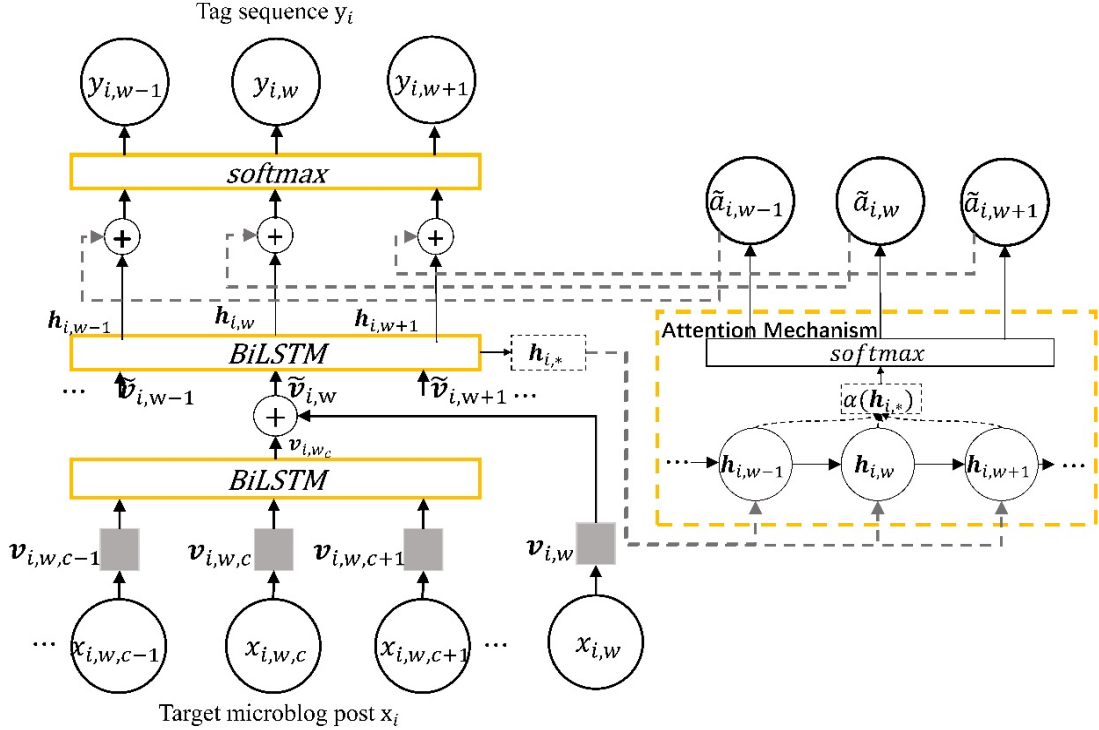

**Figure 1. Attention-mechanism-based keyphrase extraction models**

In the keyphrase extraction module, formally, given a target microblog post $x_i$ formulated as a word sequence $< x_{i,1}, x_{i,2}, \cdots, x_{i,w}, \cdots, x_{i,|x_i|} >$, where $|x_i|$ denotes the length of $x_i$, we aim to produce a tag sequence $< y_{i,1}, y_{i,2}, \cdots, y_{i,w}, \cdots, y_{i,|x_i|} >$, where $y_{i,w}$ indicates whether $x_{i,w}$ is part of a keyphrase. As shown in Figure 1, this model employs a character-level word embedding proposed by Jebbara and Cimiano (2017), which jointly uses character-level embeddings and word-level embeddings. In character-level embedding, given the character sequence $< x_{i,w,1}, x_{i,w,2}, \cdots, x_{i,w,c}, \cdots, x_{i,w,|x_{i,w}|} >$ of a word $x_{i,w}$, each character $x_{i,w,c}$ is represented by its corresponding character-level embedding $v_{i,w,c}$. The character-level embedding sequence then passes through a bidirectional long short-term memory (BiLSTM) layer (Graves & Schmidhuber, 2005), thereby obtaining a hidden state $h_{i,w}^c$ for the character sequence. The hidden state $h_{i,w}^c$ is transformed into the final character-level word embedding $v_{i,w_c}$ using a linear function. To incorporate character-level word embedding with word-level embedding, character-level word

embeddings $v_{i,w_c}$ are then concatenated with the word-level embedding $v_{i,w}$ as follows:

$$\tilde{v}_i = \{[v_{i,1}: v_{i,1_c}], [v_{i,2}: v_{i,2_c}], \cdots, [v_{i,w}: v_{i,w_c}], \cdots, [v_{i,|x_i|}: v_{i,|x_i|_c}]\} \quad (1)$$

The sequence $\tilde{v}_i$ then passes through a BiLSTM layer and thereby obtains a sequence of hidden state representations $\{h_{i,1}, h_{i,2}, \cdots, h_{i,w}, \cdots, h_{i,|x_i|}\}$ of the target microblog $x_i$. Then, $h_{i,w}$ is concatenated with $\tilde{a}_{i,w}$ to obtain the hidden state $\tilde{h}_{i,w}$ as follows:

$$\tilde{h}_{i,w} = [h_{i,w}: \tilde{a}_{i,w}] \quad (2)$$

$\tilde{a}_{i,w}$ is the normalized attention weight obtained from the following attention mechanism module:

$$\tilde{a}_{i,w} = \frac{a_{i,w}}{\sum_{|x_i|} a_{i,|x_i|}} \quad (3)$$

In the attention mechanism module, first, the hidden state $h_{i,w}$ of words in the target microblog $x_i$ is fed into the feed-forward attention module (Bahdanau et al., 2014) as follows:

$$e_{i,w} = \tanh(W_e h_{i,w} + b_e) \quad (4)$$

$$a_{i,w} = W_a e_{i,w} + b_a, \quad (5)$$

where $W_e$ and $b_e$ are parameters of function $\tanh(.)$. $W_a$ and $b_a$ are parameters of a linear function. $a_{i,w}$ is a predicted word-level attention value.

To predict the label of each word in the target microblog, the hidden state $\tilde{h}_{i,w}$ passes through a softmax layer to obtain $y_{i,w}$. Inspired by (Barrett et al., 2018), we apply two objectives: word-level loss and attention-level loss. The word-level loss minimizes the squared error between the output $y_{i,w}$ and the true label $\hat{y}_{i,w}$ as follows:

$$L_{word} = \sum_i \sum_w (y_{i,w} - \hat{y}_{i,w})^2 \quad (6)$$

Similarly, the attention-level loss is used to minimize the squared error between the attention weights $\tilde{a}_{i,w}$ and eye fixation duration $\hat{a}_{i,w}$ estimated from the OSEC as follows:

$$L_{att} = \sum_i \sum_w (\tilde{a}_{i,w} - \hat{a}_{i,w})^2 \quad (7)$$

When combined, $\lambda_{word}$ and $\lambda_{att}$ (between 0 and 1) are used to trade-off loss functions at the word-level and attention-level, respectively:

$$L = \lambda_{word} L_{word} + \lambda_{att} L_{att} \quad (8)$$

In detail, $y_{i,w}$ has five possible values $\{Single, Begin, Middle, End, Not\}$, where *Single* represents $x_{i,w}$ a one-word keyword. *Begin*, *Middle*, and *End* represent $x_{i,w}$ the first, middle, and last words of a keyphrase, respectively. *Not* represents that $x_{i,w}$ is not a keyword or not a part of a keyphrase.

### 3.2 Keyphrase Extraction Model with Human Reading Time as an External Feature

Neural network models can automatically select features from training samples. Previous studies demonstrated that manually selected features would improve the performance of neural models, such as POS (Aguilar et al., 2017), word position (Jie et al., 2017), and character position (He & Sun, 2017). In this study, we leverage human reading time as an external feature.

This model has two types of input. One is the target microblog post $x_i$ formulated as word sequence $<x_{i,1}, x_{i,2}, \cdots, x_{i,w}, \cdots, x_{i,|x_i|}>$, and the other is the TRT value of each word in the target post formulated as reading time sequence $<r_{i,1}, r_{i,2}, \cdots, r_{i,w}, , \cdots, r_{i,|x_i|}>$, where $|x_i|$ denotes the length of $x_i$. We aim to produce a tag sequence $<y_{i,1}, y_{i,2}, \cdots, y_{i,w}, \cdots, y_{i,|x_i|}>$, where $y_{i,m}$ indicates whether $x_{i,w}$ is part of a keyphrase. This model also uses the character-level word embedding proposed by Jebbara and Cimiano (2017), but we ignore this part of our architecture in the equations below.

The word sequence and human reading time sequence are fed into the embedding layer to obtain the word embedding $v_i^x$ and human reading time one-hot vector $v_i^a$. Then, $v_i^x$ and $v_i^a$ are concatenated as vector $v_i$ as follows:

$$v_i = < [v_{i,1}^x: v_{i,1}^a], [v_{i,2}^x: v_{i,2}^a], \cdots, [v_{i,w}^x: v_{i,w}^a], \cdots, [v_{i,|x_i|}^x: v_{i,|x_i|}^a] > \quad (9)$$

Figure 2 shows an example of a concatenation of $v_i^x$ and $v_i^a$ for the word 'reading.' First, we obtain the TRT value of the word 'reading.' Because the original TRT value is in continuous form and cannot be represented as a one-hot vector directly, we need to transform it into a discrete form. In this article, we propose a discrete form transformation method. First, the TRT value is rounded to the nearest hundredth. The rounded eye fixation duration is multiplied by 10 to obtain an integer value. Then, the integer form value is transformed into a one-hot vector.

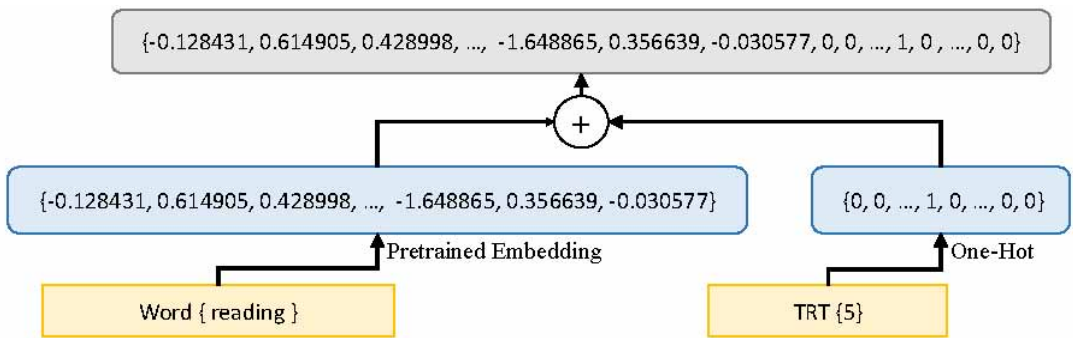

**Figure 2. Example of the concatenation of word embedding and human reading time one-hot vector**

$v_i$ passes through the BiLSTM layer to obtain the hidden state $h_{i,w}$ of each word. Then, $h_{i,w}$ passes through a softmax layer to obtain the output $y_{i,w}$ of each word. $y_{i,w}$ also has five possible values $\{Single, Begin, Middle, End, Not\}$.

## 4 Dataset

This section provides a brief introduction to the microblog datasets and OSEC used in our experiment. Section 4.1 introduces two microblog datasets. Section 4.2 gives information about the two natural-reading OSECs used in our article.

### 4.1 Keyphrase Extraction Dataset

**Election-Trec.** This dataset is based on the open-source dataset TREC2011 track[4] and the election corpus[5]. Following Q. Zhang et al. (2016), we use microblog hashtags as gold standard keyphrases. We use two rules to select microblog posts: there is only one hashtag per post, and the hashtag is inside a post. Then, we removed all the '#' symbols before keyphrase extraction. After filtering, the Election-Trec dataset contained 30, 264 Tweets. We randomly sampled 0.8, 0.1, and 0.1 for the training, development, and testing datasets. Since there are no spaces between words in hashtags, we used some strategies to segment hashtags. There are two types of hashtags in the datasets. One is the 'multi-word' type that contains both capital and lowercase letters, the other is the 'single-word' in all lowercase or all capital letters. If a hashtag is a multi-word, we segment it with two patterns. First is (capital) ∗ (lowercase) +, which represents one capital followed by one or more lowercase letters, and second is (capital) +, which represents one or more capital letters. When performing hashtag segmentation, the first pattern is used first and then the second pattern is applied. Meanwhile, we do not conduct any preprocessing if a hashtag is a single-word.

**General-Twitter Dataset.** The General-Twitter dataset[6] was released by Q. Zhang et al. (2016). This dataset does not have a development dataset, and it contains 78,760 and 33,755 Tweets in the training and testing datasets, respectively. This dataset uses hashtags as ground truth keyphrases.

We preprocessed both datasets with the Twitter NLP tool[7] for tokenization. Table 3 shows the statistical information of the two microblog datasets.

---

[4] https://trec.nist.gov/data/tweets/
[5] https://www.ccs.neu.edu/home/luwang/datasets/microblog_conversation.zip
[6] http://jkx.fudan.edu.cn/paper/data/keyphrase_dataset.tar.gz
[7] http://www.cs.cmu.edu/~ark/TweetNLP/

Table 3. Statistical information of two microblog datasets

| Dataset | Statistic Terms | # of Tweets | Tweet length | Vocab |
|---|---|---|---|---|
| **Election-Trec** | Train | 24, 210 | 19.94 | 36,018 |
| | Dev | 3, 027 | 20.00 | 9,909 |
| | Test | 3, 027 | 19.71 | 9,973 |
| **General-Twitter** | Train | 78, 760 | 13.69 | 163,115 |
| | Test | 33, 755 | 13.66 | 88,164 |

Note: Train, Dev, and Test denote the training, development, and test sets, respectively. # of Tweets: number of Tweets in the dataset. Tweet length: average word count of Tweets in the dataset. Vocab: vocabulary size of the dataset

**4.2 Eye-tracking Dataset**

**GECO.** In GECO, participants read a part of the novel 'The Mysterious Affair at Styles' by Agatha Christie. Six males and seven females whose native language was English participated in and read a total of 5,031 sentences and 9,877 unique words. There are various features in GECO, including FFD and TRT, and we use the TRT feature in this study.

**ZUCO.** ZUCO has three sub-datasets, and we used the second dataset, which is a task-free dataset that reflects natural human reading behavior. The genre of this dataset is Wikipedia, and it contains semantic relations in each sentence. To control the quality of human reading behavior, in their experiment, 68 sentences were followed by multiple-choice questions. This dataset contains 300 sentences and 2,716 unique words. Same as GECO, we use the TRT feature of ZUCO in this study.

**5 Experiments and Results**

**5.1 OSEC Preprocessing Strategy**

This section presents the OSEC preprocessing strategies used in this article. In Section 5.1.1, we introduce an initial preprocessing strategy of OSEC. In Section 5.1.2 and 5.1.3, we propose word frequency and word length regularization methods to mitigate the impact of word frequency and word length on human reading time, respectively. To solve the limited-quantity problem, in Section 5.1.4 we propose an OSEC expanding strategy to expand the word quantity in the OSEC.

**5.1.1 Initial Preprocessing Strategy of OSEC**

Because there is more than one participant in the eye-tracking experiment, we first divided the TRT values by the number of participants. Then, we calculate the average TRT values of a word in different contexts of the corpus to obtain an average TRT (AVG-TRT). For instance, the word 'aspects' occurred 14 times in the ZUCO corpus. We use the average of all the 14 TRT values to represent the eye fixation duration of the word 'aspects'.

**5.1.2. Regularization Method of Word Frequency**

To reduce the impact of word frequency on TRT, the average TRT was regularized by the frequency of words in the BNC. Before regularization, the word frequency value in the BNC was log-transformed per billion as L-BNC (Van et al., 2014). To conduct this transformation, the word frequency value in the BNC is first multiplied by 10 because the word frequency in BNC is measured per 100 million. If a word can be found in the BNC, the average TRT is multiplied by L-BNC as the frequency regularized average TRT (FRA-TRT). There are 4381 and 1081 words in GECO and ZUCO that can be found in the BNC, respectively. Otherwise, we multiple the average TRT and the average of all the L-BNC values. Then, FRA-TRT was min-max normalized to a value in the range of 0–1.

**5.1.3. Regularization Method of Word Length**

As the word length increased, the probability of fixating the word increased (Rayner & McConkie, 1976; Reichle et al., 2003). Thus, to reduce the impact of word length on TRT, inspired by Reichle et al. (2003), we propose the following equation to regularize the TRT value by word length.

$$LRA - TRT = AVG - TRT / 1.08^{Len(word)} \quad (10)$$

Where $AVG - TRT$ denotes the average TRT of a word. $Len(word)$ denotes the length of a word, which is the number of letters in a word. *1.08* is a free parameter that modulates the effects of the spatial disparity between each word's letters and the fixation location (Reichle et al., 2003). The word length regularized average TRT (LRA-TRT) was then min-max normalized to a value in the range of 0–1.

**5.1.4. OSEC Expanding Strategy**

As we can see from Table 2, there are a large number of words in the target datasets that cannot be found in OSEC. We call these words Out-OSEC words. Otherwise, we use In-OSEC words to denote a word in the target dataset that can be found in the OSEC. We propose an expanding strategy to expand the word quantity in OSEC. First, we calculate the cosine similarity between an Out-OSEC word and an In-OSEC word in the target microblog dataset. There are two possible situations when calculating the similarity between two words. First, if both words have corresponding pre-trained word embeddings, we calculate the cosine similarity of the two pre-trained word embeddings. The second is that if one of the words does not have a pre-trained word embedding, we calculate the cosine similarity between the two character-level word embeddings. The character-level word embedding is trained by the BiLSTM model introduced in Section 5.2.2. We then select the top 10 In-OSEC words having the highest cosine similarity value with the Out-OSEC word. These words are used to represent the human reading time of the Out-OSEC word. As shown in the equation (see below), $\mathrm{E-TRT}$ denotes the expanded TRT value of the Out-OSEC word. $AVG-TRT_i$ denotes the TRT of the top $i$ In-OSEC word. $\text{Weight}_i$ is the weight of the top $i$ In-OSEC word and is calculated by $\text{Weight}_i = i \times 1/\sum_{i=1}^{10} i$.

$$\mathrm{E-TRT} = \sum_{i=1}^{10} \text{Weight}_i \times AVG - TRT_i \qquad (11)$$

## 5.2 Experimental Setting

### 5.2.1. Implementation details

In the training, we chose a BiLSTM with 300 dimensions, and $\lambda_{word}$ and $\lambda_{att}$ were set to 0.7 and 0.3, respectively. Hyperparameters were set under the best performance. Because the General-Twitter dataset does not have a development dataset, it had the same hyperparameters as the Election-Trec dataset when training. The epoch was set to 10 and 5 when training with Election-Trec dataset and General-Twitter dataset, respectively. We used the RMSprop optimizer (Graves, 2013). We initialized target posts by embeddings pre-trained on 99 M Tweets with 27 B tokens and 4.6 M words in the vocabulary. All the models were trained five times, and we used the average macro P, R, and $F_1$ of the five models to analyze the performance of the models.

### 5.2.2. Baseline Models

We compared our models with two kinds of neural network models, namely the BiLSTM model and the BiLSTM model with the attention mechanism (Att-BiLSTM model).

**BiLSTM model.** This model is a sequence labeling model constructed by character-level word embedding and the BiLSTM layer.

**Att-BiLSTM model**. This model is a BiLSTM model with an attention mechanism. The output of the attention mechanism is not modified by human reading time.

### 5.2.3. Our Models

**HA-BiLSTM-Ex model.** This model is a BiLSTM model with an attention mechanism whose output is modified using human reading time (see Section 3.1). The human reading time used in this model is collected from expanded OSEC, and it is regularized by word frequency and word length regularization strategies.

**F-BiLSTM-Ex model.** This model is a BiLSTM model with human reading time as the external feature (see Section 3.2). The human reading time is regularized by word frequency and word length regularization strategies, which is collected from expanded OSEC. The human reading time is then transformed into a one-hot vector.

### 5.3 Result

In this section, we first compare the overall performance of the baseline models and our proposed models. Then, we conduct three ablation experiments to analyze whether the word frequency regularization method, word length regularization method, and the OSEC expanding strategy will improve the performance of keyphrase extraction models. We also conducted qualitative experiments to analyze the impact of human reading time in our proposed neural network models.

#### 5.3.1. Overall performance

Table 4. Macro-averaged precision, recall, and $F_1$ scores of the baselines and proposed models on the Election-Trec and General-Twitter datasets (%)

| Models | | Dataset | Election-Trec | | | General-Twitter | | |
|---|---|---|---|---|---|---|---|---|
| | | | P | R | $F_1$ | P | R | $F_1$ |
| Baseline | | BiLSTM | 74.84 | 67.57 | 71.02 | <u>80.81</u> | 74.62 | 77.59 |
| | | Att-BiLSTM | <u>75.06</u> | <u>67.96</u> | <u>71.32</u> | 80.56 | <u>75.29</u> | <u>77.83</u> |
| Our Model | GECO | HA-BiLSTM-Ex | **75.95** | 68.56 | **72.06** | 81.47 | <u>75.65</u> | **78.46** |
| | | F-BiLSTM-Ex | <u>75.60</u> | **68.73** | 72.00 | <u>81.59</u> | 75.55 | 78.45 |
| | ZUCO | HA-BiLSTM-Ex | <u>75.85</u> | 68.37 | 71.91 | 80.00 | **76.64** | <u>78.28</u> |
| | | F-BiLSTM-Ex | 74.59 | <u>69.58</u> | <u>71.99</u> | **83.13** | 73.74 | 78.16 |

Note: Bold font and underline indicate the best score in each column and each block, respectively.

We compared the performance of our proposed models with those of two baseline models on two microblog datasets. Our proposed models contain two categories of models with human reading time, including the HA-BiLSTM-Ex model and F-BiLSTM-Ex model. The two baseline models include the BiLSTM and Att-BiLSTM models, which do not use human reading time.

Table 4 shows the results of these models. From Table 4, we can make two observations:

- ***The attention mechanism modified by human reading time can improve the performance of neural network keyphrase extraction models.*** As shown in Table 4, all $F_1$ values of the Att-BiLSTM models are higher than those of the BiLSTM models. This indicates that the attention mechanism can improve the performance of sequence-labeling keyphrase extraction models. Moreover, the HA-BiLSTM-Ex models yield a higher $F_1$ score than the Att-BiLSTM models. In contrast to Att-BiLSTM, in the HA-BiLSTM-Ex model, human reading time is used to modify the output of the attention mechanism. Thus, it indicates that the attempt to use human reading time as the ground truth of the output of attention mechanisms is feasible.

- ***The human reading time external feature can determine salient words.*** As shown in Table 4, all $F_1$ values of F-BiLSTM-Ex are higher than those of the baseline models. Since F-BiLSTM-Ex uses human reading time as an external feature, it indicates that human reading time is a useful feature to distinguish the importance of words in a microblog.

Moreover, to analyze whether the genre of OSEC has an impact on the performance of keyphrase extraction models, we used two OSECs in different genres, that is, GECO and ZUCO. The genres of GECO and ZUCO are novel and Wikipedia, respectively. Table 4 shows the performance of the models with these two OSECs. From Table 4, we find the following:

- ***Eye-tracking corpora in different genres can help models extract salient phrases.*** As shown in Table 4, models with eye fixation durations extracted from both GECO and ZUCO yielded better performance than the two basic models. This indicates that both OSECs can improve the performance of keyphrase extraction on target datasets.

- ***The eye fixation duration collected from GECO is more efficient than that from ZUCO.*** Although our proposed models yielded better performance than the basic models, the $F_1$ score of models with eye fixation durations collected from GECO was higher than that of ZUCO. There are two reasons for this phenomenon. First, the word quantity in GECO is larger than ZUCO, and hence more words in the target dataset have corresponding eye fixation durations. Second, the genre of GECO is more similar to microblog posts than

ZUCO. This is because the genre of GECO is novel, which is close to common human expression, while the genre of ZUCO is Wikipedia, which contains semantic relations.

Although the experimental results show that the human reading time obtained from GECO and ZUCO can help indicate salient phrases from Tweet, previous studies state that the human reading time collected from a specific eye-tracking corpus is tied to the words in the context, and cannot be directly used as the general measure of reading time of that word (Dirix et al., 2019). Thus, we conduct a further experiment to analyze whether the human reading time obtained from a specific eye-tracking corpus is efficient in the keyphrase extraction task. In the experiment, with the assumption that the average reading time of a word in many contexts could be closer to a pure reading time (Dirix et al., 2019), we combined two OSECs, i.e., GECO and ZUCO. Then, the performance of keyphrase extraction models using human reading time obtained from specific OSECs is compared with that using human reading time collected from combined OSEC. The OSECs combination strategy and experimental results are shown in Section 5.3.2.

**5.3.2. Experiment with Combined Eye-tracking Corpus**

In this section, we first describe the OSEC combination strategy. Then, we analyze experimental results from two aspects. One is whether the human reading time obtained from a specific OSEC is efficient in keyphrase extraction; the other is whether the combined OSEC can help improve the performance of keyphrase extraction models.

The OSEC combination strategy has two steps. First, we use the initial preprocessing strategy in Section 5.1.1 to process the eye fixation duration in GECO and ZUCO. Then, we collect the co-occurring words in both OSECs and calculate the average of the eye fixation duration of these words in two OSECs. The average value of the eye fixation duration is utilized for representing the human reading time of these words.

In the experiment, we employ the HA-BiLSTM-EX and F-BiLSTM-EX models, and compare the performance of models using human reading time collected from specific OSECs with the performance of models using human reading time collected from combined OSEC. The experimental results of models using specific OSECs and combined OSEC are shown in Table 4 and Table 5, respectively. From these two Tables, we have two observations. First, **specific OSECs are efficient in improving the performance of keyphrase extraction models.** The F1 scores of models with human reading time from GECO shown in Table 4 are higher than the scores of models with human reading time collected from combined OSEC shown in Table 5. The reason for this phenomenon is that the size of the co-occurring words in two OSECs is small (575 words), and a large number of words in the target Twitter dataset cannot find the corresponding eye fixation duration from the combined OSEC. What's more, since the genre of GECO is novel, which is close to the common human expression on Twitter, this observation also proves the conclusion that OSECs in genres that are close to the target dataset is more efficient. Second, **the human reading time collected from the combined OSEC can help identify salient phrases from Tweets.** By comparing Table 4 and Table 5, we found that using eye fixation duration obtained from combined OSEC, our proposed models yield better F1 scores than the baseline models. This proves that our proposed OSEC combination strategy is reasonable.

**Table 5. Macro-averaged precision, recall, and $F_1$ scores of the proposed models with human reading time collected from combined OSEC on the Election-Trec and General-Twitter datasets (%)**

| Dataset<br>Models | Election-Trec | | | General-Twitter | | |
|---|---|---|---|---|---|---|
| | P | R | F1 | P | R | F1 |
| HA-BiLSTM-Ex | 75.85 | **68.37** | 71.91 | <u>81.39</u> | 75.45 | **78.31** |
| F-BiLSTM-Ex | **76.08** | 68.20 | **71.92** | 80.66 | <u>76.01</u> | 78.26 |

Note: Bold font and underline indicate the best score in each column and each block, respectively

### 5.3.3. Ablation Experiment

As mentioned in Section 1, OSEC has three drawbacks: the word quantity in OSEC is limited, and the word frequency and word length have a negative impact on eye-tracking values. We propose a similarity-based OSEC expanding strategy, a frequency regularization strategy, and a length regularization strategy to mitigate these three problems, respectively. In this section, we analyze whether these three strategies are effective.

**(1) Impact of the OSEC Expanding Strategy**

As mentioned in Section 1, because the number of words in OSEC is limited, there are many words in the target microblog datasets that cannot be found in OSEC. Thus, we propose an OSEC expanding strategy introduced in Section 5.1.4 to expand the word quantity in OSEC. This section aims to analyze whether this strategy is effective in mitigating the limited-quantity problem. We compared the P, R, and $F_1$ scores of two different kinds of models, that is, models using expanded OSEC (HA-BiLSTM-Ex and F-BiLSTM-Ex) and models using unexpanded OSEC (HA-BiLSTM and F-BiLSTM). In HA-BiLSTM and F-BiLSTM, if a word in the target microblog dataset cannot be found in the OSEC corpus, we assign a mean average of human reading time to the word. The results of HA-BiLSTM and F-BiLSTM are shown in Table 6, and the results of HA-BiLSTM-Ex and F-BiLSTM-Ex are shown in Table 4. From these two Tables, we found that ***both the HA-BiLSTM-Ex and F-BiLSTM-Ex models yielded better $F_1$ scores than the HA-BiLSTM and F-BiLSTM models, regardless of the genre of OSEC***. This observation indicates that the OSEC expanding strategy can effectively alleviate the limited-quantity problem and thereby improve the performance of keyphrase extraction models.

**Table 6. Average macro precision, recall, and $F_1$ scores of the HA-BiLSTM and F-BiLSTM models on the Election-Trec and General-Twitter datasets (%)**

| Models | Dataset | Election-Trec | | | General-Twitter | | |
|---|---|---|---|---|---|---|---|
| | | P | R | $F_1$ | P | R | $F_1$ |
| GECO | HA-BiLSTM | <u>75.74</u> | 68.26 | <u>71.79</u> | 81.40 | <u>75.32</u> | **<u>78.24</u>** |
| | F-BiLSTM | 75.43 | <u>68.41</u> | 71.74 | **<u>81.98</u>** | 74.76 | 78.20 |
| ZUCO | HA-BiLSTM | **<u>75.81</u>** | 68.30 | **<u>71.85</u>** | <u>80.91</u> | 75.61 | <u>78.17</u> |
| | F-BiLSTM | 75.37 | **<u>68.62</u>** | 71.83 | 80.28 | **<u>76.02</u>** | 78.09 |

Note: HA-BiLSTM is the model with an attention mechanism whose outputs are regularized with human reading time, and F-BiLSTM is the model with the human reading time as an external feature. The OSEC used in these models was not expanded by the expanding strategy. Bold font and underline indicate the best score in each column and each block, respectively.

**(2) Impact of Word Frequency Regularization Strategy**

As mentioned in Section 1, to reduce the impact of word frequency on human reading time, we used the word frequency extracted from the BNC corpus to regularize the human reading time. To analyze whether this word frequency regularization strategy is efficient, we initially constructed models with human reading time that is not regularized with word frequency and word length (HA-BiLSTM-N and F-BiLSTM-N). Then, we compared these two models with HA-BiLSTM-Fre and F-BiLSTM-

Fre, in which the human reading time is regularized with word frequency. The results of the HA-BiLSTM-N, F-BiLSTM-N, HA-BiLSTM-Fre, and F-BiLSTM-Fre models are shown in Table 7. From Table 7, we found that *when using GECO and ZUCO, all the HA-BiLSTM-Fre and F-BiLSTM-Fre models yielded better F1 scores than the HA-BiLSTM-N and F-BiLSTM-N models.* This indicates that the word frequency regularization strategy plays an important role in GECO and ZUCO.

**(3) Impact of Word Length Regularization Strategy**

To reduce the impact of word length on human reading time, we used the word length to regularize the reading time. To analyze whether this word length regularization strategy is efficient, we compared HA-BiLSTM-N and F-BiLSTM-N with HA-BiLSTM-Len and F-BiLSTM-Len, in which the human reading time is regularized with word length. The results of the HA-BiLSTM-N, F-BiLSTM-N, HA-BiLSTM-Len, and F-BiLSTM-Len models are shown in Table 7. From Table 7, we have two observations. First, *when using ZUCO, all the HA-BiLSTM-Len and F-BiLSTM-Len models yielded better F1 scores than the HA-BiLSTM-N and F-BiLSTM-N models.* This indicates that the word length regularization strategy plays an important role in ZUCO. Second, *when using GECO, all the HA-BiLSTM-Len models yielded better F1 scores than the HA-BiLSTM-N in the Election-Trec and General-Twitter dataset, while the $F_1$ scores of the F-BiLSTM-N models in both target dataset were higher than those of the F-BiLSTM-Len models.* This indicates that when eye fixation durations in GECO are used as an external feature, the word length regularization strategy does not have a positive effect on it. In this scenario, regularizing other factors, such as verb complexity and lexical ambiguity, may play an important role in the performance of keyphrase extraction models.

**Table 7. Average macro precision, recall, and $F_1$ scores of the HA-BiLSTM-N, F-BiLSTM-N, HA-BiLSTM-Fre, F-BiLSTM-Fre, HA-BiLSTM-Len, and F-BiLSTM-Len models on the Election-Trec and General-Twitter datasets (%)**

| Dataset / Models | | Election-Trec | | | General-Twitter | | |
|---|---|---|---|---|---|---|---|
| | | **P** | **R** | **$F_1$** | **P** | **R** | **$F_1$** |
| GECO | HA-BiLSTM-N | 76.38 | 67.27 | 71.53 | 81.67 | 74.50 | 77.92 |
| | F-BiLSTM-N | 76.57 | 66.90 | 71.39 | 81.19 | 75.20 | 78.08 |
|  | HA-BiLSTM-Fre | <u>76.18</u> | 67.71 | <u>71.67</u> | 80.36 | **<u>76.04</u>** | 78.13 |
|  | F-BiLSTM-Fre | 75.35 | <u>68.14</u> | 71.53 | <u>82.03</u> | 75.21 | **<u>78.47</u>** |
|  | HA-BiLSTM-Len | <u>76.55</u> | <u>67.81</u> | <u>71.90</u> | 80.56 | <u>75.78</u> | <u>78.10</u> |
|  | F-BiLSTM-Len | 76.45 | 66.93 | 71.35 | <u>80.83</u> | 75.28 | 77.95 |
| ZUCO | HA-BiLSTM-N | 75.41 | <u>67.93</u> | 71.47 | 81.03 | 75.28 | <u>78.04</u> |
|  | F-BiLSTM-N | <u>76.01</u> | 67.65 | <u>71.56</u> | 80.82 | 75.41 | 78.02 |
|  | HA-BiLSTM-Fre | 75.18 | **68.98** | **71.94** | 81.98 | <u>74.74</u> | 78.19 |
|  | F-BiLSTM-Fre | 75.01 | 68.35 | 71.50 | **82.34** | 74.58 | <u>78.27</u> |
|  | HA-BiLSTM-Len | <u>**76.61**</u> | 67.45 | <u>71.73</u> | 82.18 | 74.50 | <u>78.15</u> |
|  | F-BiLSTM-Len | 75.79 | <u>67.73</u> | 71.52 | 80.97 | <u>75.35</u> | 78.05 |

Note: HA-BiLSTM-N is the model with an attention mechanism whose output is modified by human reading time. F-BiLSTM-N is the model with the human reading time as an external feature. The human reading time used in these models is not regularized by word frequency and is not expanded by the word expanding strategy. The human reading time used in HA-BiLSTM-Fre and F-BiLSTM-Fre models is regularized by word frequency but is not expanded by the word expanding strategy. The human reading time used in HA-BiLSTM-Len and F-BiLSTM-Len models is regularized by word length but is not expanded by the word expanding strategy. Bold font and underline indicate the best score in each column and each block, respectively.

**5.3.4. Qualitative Analysis of Human Reading Time in Neural Models**

The results in Section 5.3.1 show that our proposed neural models yielded better $F_1$ scores than models without human reading time. This section aims to analyze the role of human reading time in neural models.

**(1) Role of human reading time in the attention mechanism**

To analyze why human reading time has a positive effect on attention-mechanism-based neural models, this section aims to analyze the difference between the outputs of the attention mechanism of HA-BiLSTM-Ex and Att-BiLSTM.

First, we build a sub-microblog dataset, from which HA-BiLSTM-Ex models extract the right keyphrases, while the words extracted by Att-BiLSTM are different from the ground truth.

$$Sub_{test} = \{x_j | r_j^{HA-BiLSTM-Ex} = y_j \cap r_j^{Att-BiLSTM} \neq y_j, 0 \leq j \leq n\}, \quad (12)$$

where $x_j$ is the $j_{th}$ microblog, and $y_j$ is the ground truth keyphrase of $x_j$. $r_j^{HA-BiLSTM-Ex}$ and $r_j^{Att-BiLSTM}$ are the keyphrases predicted by the HA-BiLSTM-Ex and Att-BiLSTM models, respectively. There are 205 and 1,203 Tweets in the sub-microblog datasets of the Election-Trec and General-Twitter datasets, respectively. Then, we collected the outputs of the attention mechanism of the HA-BiLSTM-Ex and Att-BiLSTM models on the ground truth keyphrases and compared the distributions of these two outputs. Because the outputs of the attention mechanism of different models cannot be compared directly, we use the rank order to represent them. Here, output values are sorted in descending order, by which the bigger values are ranked higher. If a ground-truth keyphrase consists of more than one word, we use the mean average value of all the words in the keyphrase.

Figure 3 and 4 show the distribution of all rank orders of the Election-Trec and General-Twitter sub-datasets, respectively. The triangles denote the rank order of Att-BiLSTM, and the circles denote the rank order of HA-BiLSTM-Ex. From Figure 3, in the Election-Trec dataset, most rank orders of output values of HA-BiLSTM-Ex are lower than 30, while most rank orders of Att-BiLSTM are greater than 40. From Figure 4, in the General-Twitter dataset, most rank orders of HA-BiLSTM-E are lower than 20, while most rank orders of Att-BiLSTM are higher than 20. The above observations indicate that ***most outputs of attention mechanisms in HA-BiLSTM-Ex are larger than those in Att-BiLSTM.*** Hence, using human reading time as the ground truth can effectively modify the attention mechanism and thereby increase the output values of the attention mechanism on ground truth keyphrases.

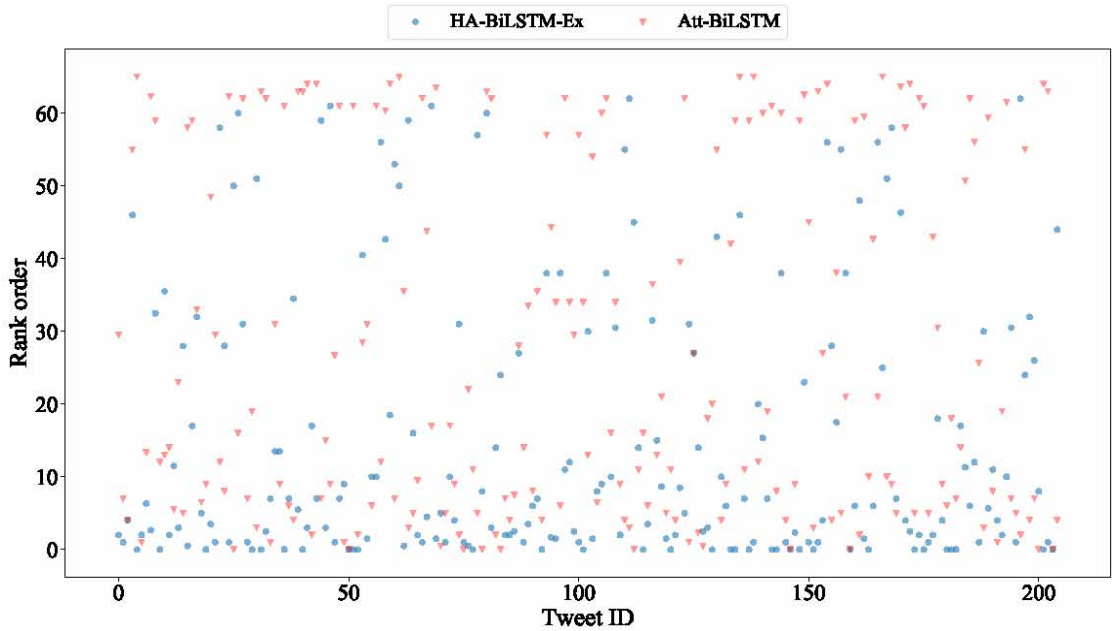

**Figure 3. Distribution of rank orders on Election-Trec dataset**

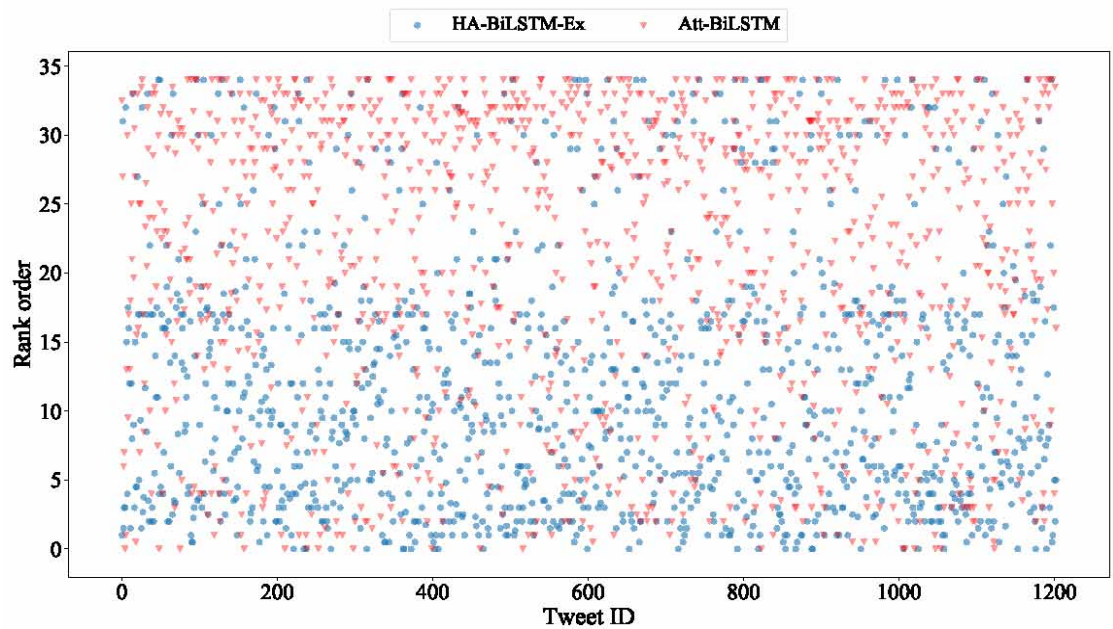

**Figure 4. Distribution of rank orders on the General-Twitter dataset**

**(2) Role of the human reading time as an external feature**

Intuitively, external features provide information that cannot be learned from the training dataset, which affects the generalization ability of neural models. To analyze whether human reading time external features will improve the generalization ability of keyphrase extraction models, inspired by Fu et al. (2020), we constructed a sub-set of the testing dataset in which the ground truth keyphrases do not exist in the training dataset. The intuition behind this is that we want the model to predict a keyphrase without prior information of the keyphrase itself. There are 1,185 Tweets and 4,405 Tweets in the sub-set of Election-Trec and General-Twitter testing datasets, respectively.

We compared the $F_1$ scores of the F-BiLSTM-Ex and BiLSTM models on the sub-set, as shown in Table 8. Based on this Table, there are two observations. ***First, the $F_1$ score of the sub-set is smaller than that of the whole dataset.*** This indicates that it is difficult to extract keyphrases that do not exist in the training dataset. Second, ***using human reading time as an external feature, the $F_1$ score of F-BiLSTM-Ex on two sub-datasets is better than that of BiLSTM***. Hence, human reading time can improve the generalization ability of keyphrase extraction models.

**Table 8. Performance of the BiLSTM and F-BiLSTM-Ex models on two sub-datasets**

| Models | Dataset | Election-Trec | | | General-Twitter | | |
|---|---|---|---|---|---|---|---|
| | | P | R | $F_1$ | P | R | $F_1$ |
| BiLSTM | | 59.98 | 55.01 | 57.39 | 38.00 | 30.60 | 33.90 |
| F-BiLSTM-Ex | GECO | **66.57** | **55.26** | 60.39 | **42.58** | 30.00 | 35.20 |
| | ZUCO | 65.59 | 53.85 | 59.14 | 42.49 | **32.52** | **36.85** |

Note: Bold font and underline indicate the best score in each column and each block, respectively.

## 6 Conclusion and Future Work

In this paper, we proposed two types of human-reading-time-based keyphrase extraction models. One is a neural model with an attention mechanism whose outputs are regularized by human reading time, and the other is a neural model with the human reading time as an external feature. Both proposed models yield better performance than the baseline models without human reading time. The reading times were extracted from OSEC. We analyzed three main problems in the OSEC and proposed a corresponding strategy to mitigate these drawbacks. First, we used two OSECs in different genres and found that OSECs in genres that are close to common human expression are more efficient. Second, the regularization strategy can mitigate the impact of word frequency and word length on human reading time to a great extent. Third, the OSEC expanding strategy had a positive effect on the performance of keyphrase extraction.

For OSEC, there are still many areas to explore. First, learned from Section 5.3.2, although the human reading time collected from the combined OSEC could improve the performance of keyphrase extraction models; this improvement is not obvious because of the small quantity of co-occurring words in different OSEC. Thus, in the future, proposing OSEC combination strategies to alleviate the limited-word problem is helpful. Second, besides the TRT, there are other features of OSEC. Hence, in the future, it will be useful to analyze the performance of other OSEC features on the keyphrase extraction task. Moreover, eye fixation duration is influenced by other factors besides word frequency and word length, such as verb complexity and lexical ambiguity. Therefore, proposing regularization strategies to alleviate the impact of other influencing factors is also significant. In addition to microblogs, our proposed models can be used for extracting keyphrase models from documents in other genres, such as the news and academic papers. Furthermore, the two models proposed in this article can be easily merged into one model for further studies.

**Acknowledgments**

This study is supported by the National Natural Science Foundation of China (Grant No.72074113), Science Fund for Creative Research Group of the National Natural Science Foundation of China (Grant No. 71921002)